\documentclass[conference]{IEEEtran}
\IEEEoverridecommandlockouts
\usepackage{cite}
\usepackage{amsmath,amssymb,amsfonts}
\usepackage{algorithmic}
\usepackage{graphicx}
\usepackage{textcomp}
\usepackage{xcolor}
\usepackage{subfigure}
\usepackage[ruled,vlined,noresetcount]{algorithm2e}
\def\BibTeX{{\rm B\kern-.05em{\sc i\kern-.025em b}\kern-.08em
    T\kern-.1667em\lower.7ex\hbox{E}\kern-.125emX}}

\begin{document}

\title{Surface Damage Detection using a Combination of Convolutional Neural Network and Artificial Neural Neural\\
}

\author{\IEEEauthorblockN{Alice Yi Yang}
\IEEEauthorblockA{\textit{School of Electrical and Information Engineering} \\
\textit{University of the Witwatersrand}\\
Johannesburg, South Africa \\
yangalice8@gmail.com}
\and
\IEEEauthorblockN{Ling Cheng}
\IEEEauthorblockA{\textit{School of Electrical and Information Engineering} \\
\textit{University of the Witwatersrand}\\
Johannesburg, South Africa \\
ling.cheng@wits.ac.za}
}

\maketitle

\begin{abstract}
Surface damage on concrete is important as the damage can affect the structural integrity of the structure. This paper proposes a surface damage detection scheme using Convolutional Neural Network (CNN) and Artificial Neural Network (ANN). The CNN classifies given input images into two categories: positive and negative. The positive category is where the surface damage is present within the image, otherwise the image is classified as negative. This is an image-based classification. The ANN accepts image inputs that have been classified as positive by the ANN. This reduces the number of images that are further processed by the ANN. The ANN performs feature-based classification, in which the features are extracted from the detected edges within the image. The edges are detected using Canny edge detection. A total of 19 features are extracted from the detected edges. These features are inputs into the ANN. The purpose of the ANN is to highlight only the positive damaged edges within the image. The CNN achieves an accuracy of 80.7\% for image classification and the ANN achieves an accuracy of 98.1\% for surface detection. The decreased accuracy in the CNN is due to the false positive detection, however false positives are tolerated whereas false negatives are not. The false negative detection for both CNN and ANN in the two-step scheme are 0\%.
\end{abstract}

\begin{IEEEkeywords}
Artificial Neural Network, Convolutional Neural Network, Feature Extraction, Classification, Machine Learning, Principal Component Analysis, Surface Damage Detection
\end{IEEEkeywords}

\section{Introduction}
Concrete degradation is a result of elemental reactions that the concrete is exposed to, such as fire, chemical, physical damage and calcium leaching \cite{glasser2008durability}. The concrete damage poses a danger to its surrounding environment as it presents structural damages. Severe concrete damages can result to the collapsing of structural buildings. There are many ways to measure concrete degradation, which involves considering multiple parameters of the concrete, namely, concrete pH, concentration of solution, physical state, and temperature rate of replenishment \cite{de2007evaluation}. However, the most basic form of concrete evaluation is through visualization. In this paper, an scheme for surface damage detection for image data using a combination of Convolutional Neural Network (CNN) and Artificial Neural Network (ANN). The objective of the CNN is to classify positive and negative surface damage images, whereas the ANN classifies and highlights the identified edges in the image as positive or negative damage. The paper is structured such that Section This paper is structured such that Section \ref{Section: Literature} provides literature on existing solutions for concrete damage
detection, whilst Section \ref{Section: Background} gives a brief background on the neural networks utilised in the scheme. Section \ref{Section: System Overview} describes the main components the system is constructed with, as well as the overview of both the training and the execution of the system. A detailed description of the proposed scheme is presented in Section \ref{Section: Methodology}, which includes the training and execution procedure of the two-step scheme. Section \ref{Section: Results and Analysis} illustrates the results and analysis obtained from the two-step scheme. Finally, in Section \ref{Section: Future Improvements} addresses the future improvements to the two-step scheme.

\section{Literature} \label{Section: Literature}
The authors of \cite{cai2018methods} adopted computer vision technology to improve the efficiency of crack detection for concrete bridge structures. The proposed system consists of three components: high magnification image acquisition system, a 2D electric cradle head, and a laser ranging system. The proposed system maps observed co-ordinates with marking points and based on the marking points the cracks are measured in the spatial location. The system performs image preprocessing such that the highest resolution image is processed with defined model of median filtering. The preprocessing reduces the noise within the image. The locating of the cracks are performed based on the image coordinate and the marking points. The authors show that the results can automatically measure cracks within 16 seconds at a distance of 100 meters with a deviation within $\pm 7^{\circ}$.

Digital image processing techniques and algorithms are proposed by the authors of \cite{li2015study} for crack detection. The techniques include image pre-processing, crack extraction, and crack connection algorithm. The image pre-processing consists of transforming the image from colour to greyscale. De-noising is applied to the image to eliminate noise by employing a spatial filter. The resultant image is enhanced to improve the visual effect and sharpness of the image. The extraction algorithm employs multiple image edge detection algorithms: Robert operator, Prewitt operator, Laplacian operator, Canny factor, and Sobel factor. The detected edges are further processed using Mathematical Morphology (MM), which is an analysis tool for geometrical structures. The crack detection algorithm is performed based on Wavelet transform by determining the different spatial resolution at different frequencies and direction characteristics. A crack connection algorithm is employed to check the integrity of the extracted crack.

Li, W. et al. presents a crack detection method for track slabs in complex environments \cite{li2018track}. The detection methods involves histogram equalisation to reduce illumination effects generated from capturing the data during night time with limited lighting sources. Additionally, an improved Canny edge detection algorithm for continuous crack profile detection is included. The authors further propose an algorithm based on the contour of fracture shape which is used to extract fracture characteristics to achieve accurate crack location. The authors achieved an accuracy of 87.3\% for crack detection using the proposed method.

Cho, H. et al. describes an image-based methodology for structural cracks detection in concrete \cite{cho2018image}. The image-based methodology consists of five distinct steps, namely, crack width transform, aspect ratio filtering, crack region search, hole filling, and relative thresholding. The implementation of the proposed methodology achieved an accuracy of 82\% for crack identification and classification.

\section{Background} \label{Section: Background}
\subsection{Convolutional Neural Network}
A Convolutional Neural Network (CNN) is a supervised neural network that is commonly utilised for image classification \cite{howard2013some, li2014medical}. It employs the mathematical operation called convolution for automated feature extraction. The CNN performs its image classification through the process of summarizing the extracted image features. The summarized information will indicate the category of the image, based on the training of the CNN. Since the CNN is a supervised neural network, it is trained with labelled data. 

\subsection{Artificial Neural Network}
An Artificial Neural Network (ANN) is a simple neural network which consists of layers of nodes \cite{yegnanarayana2009artificial}. Similar to CNNs, it can be used for classification and decision making. The ANN diffs from the CNN as it does not perform automated feature extraction, rather it accepts any one-dimensional input. These inputs can be in the form of pixels or features, which is all dependent on the data. Similar to the CNN, ANN also requires supervised learning, hence it is trained with labelled data.

\section{System Overview} \label{Section: System Overview}
\begin{figure}[h!]
	\centering
	\includegraphics[scale=0.45]{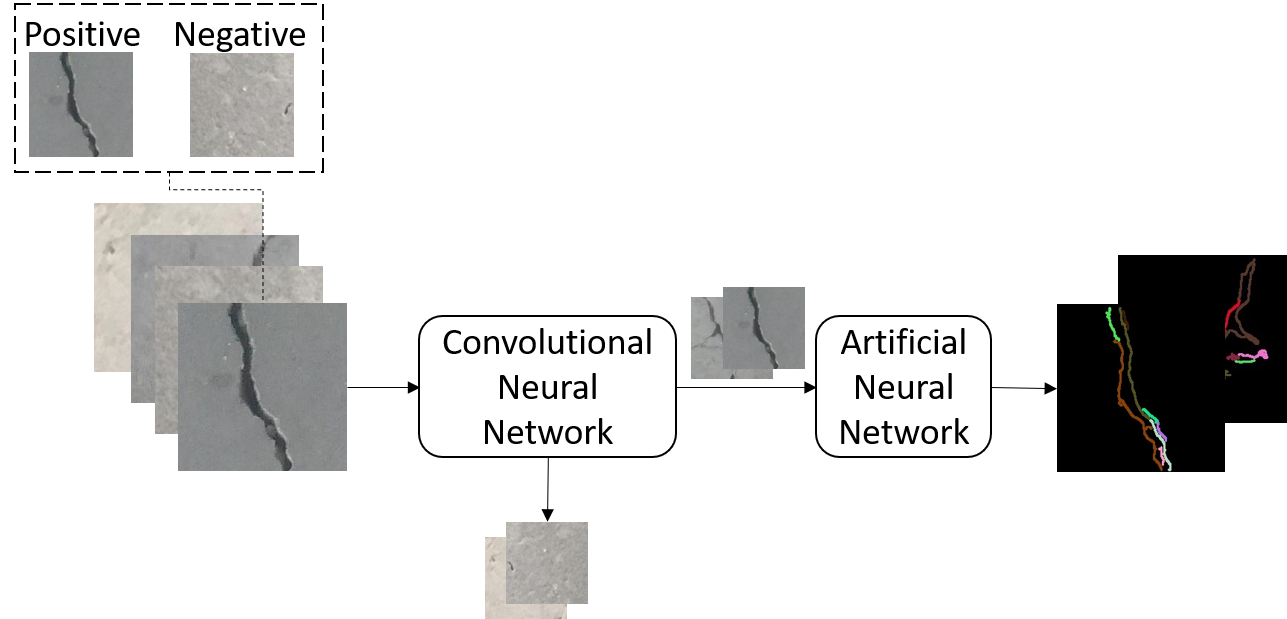}
	\caption{Diagram illustrating the overview of the scheme.}
	\label{fig: system_overview}
\end{figure}

\begin{figure}[h!]
	\centering	
	\includegraphics[scale=0.8]{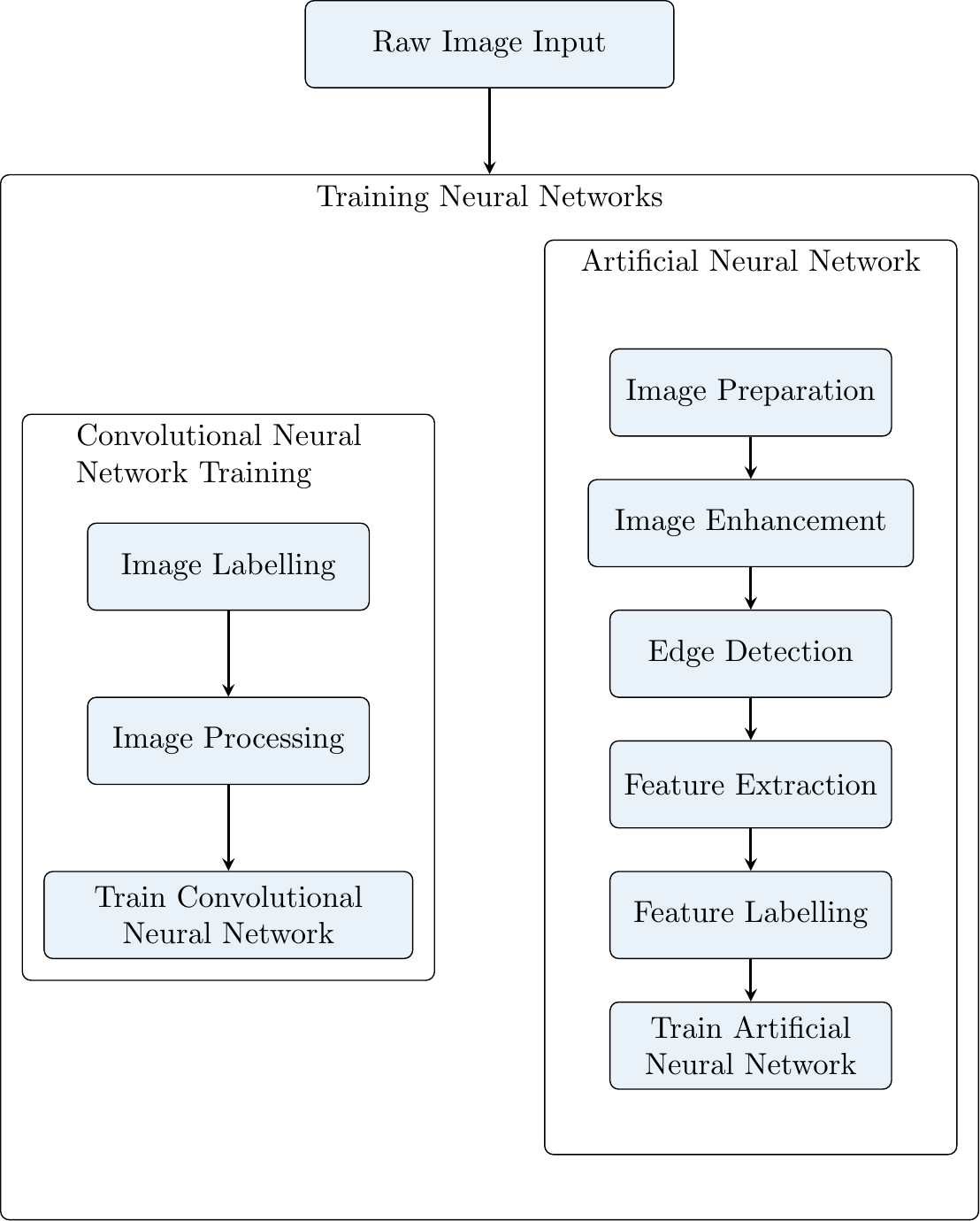}
	\caption{Block diagram illustrating the overview system for training the both neural network components}
	\label{Figure: Training System}
\end{figure}

\begin{figure}[h!]
	\centering	
	\includegraphics[scale=0.7]{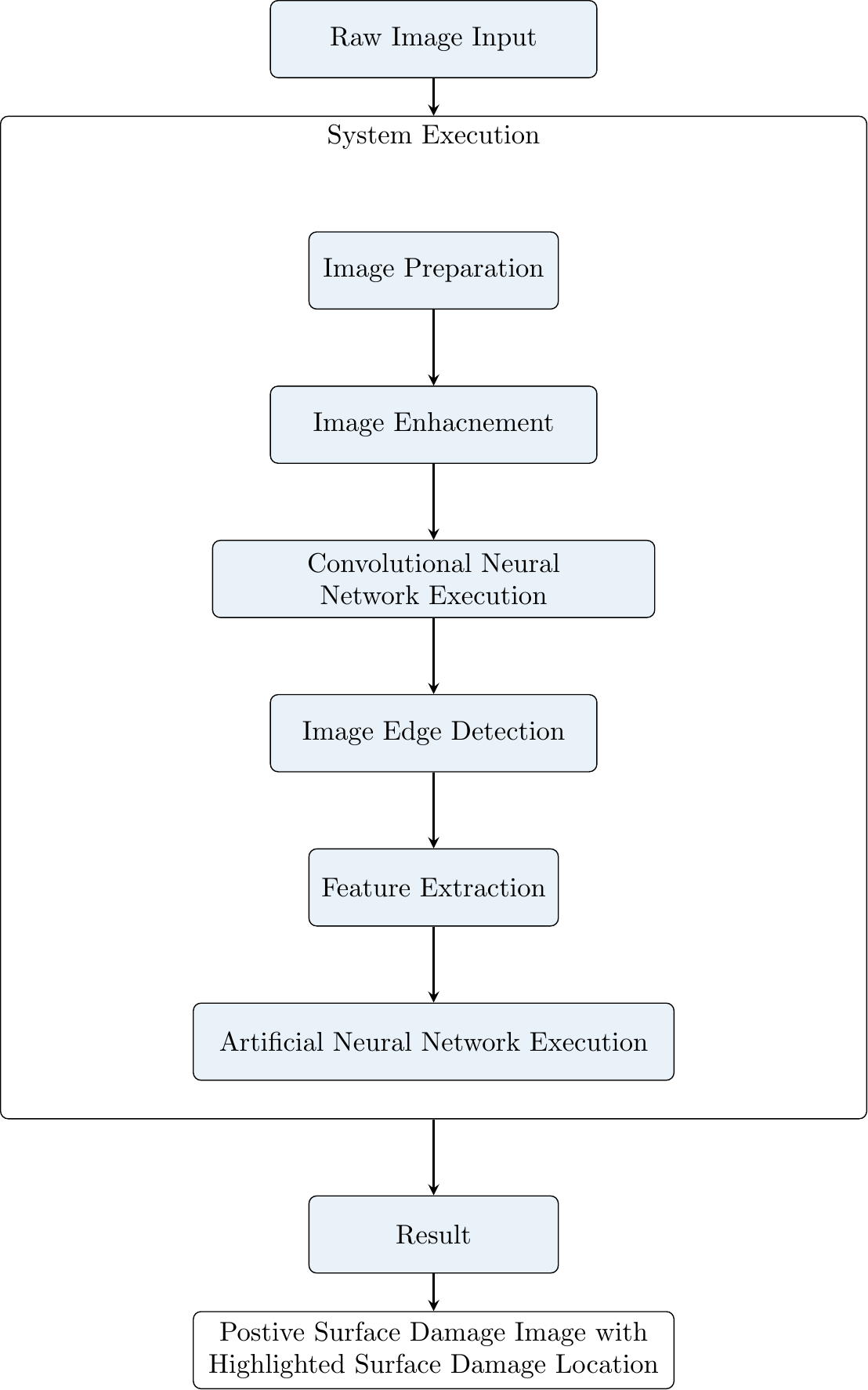}
	\caption{Block diagram illustrating the overview system for executing both neural network components for surface damage detection}
	\label{Figure: Execution System}
\end{figure}

The algorithm proposed consists of two neural network components, namely, CNN and ANN. The CNN is used to classify positive from negative surface damage images, whilst the ANN highlights and locates the surface damage within the image. Both CNN and ANN require training, as they are supervised neural networks. An overview of the two-step scheme is shown in Figure \ref{fig: system_overview}. The CNN acts as a filter to only allow images with surface damage present to be further processed by the ANN. Figure \ref{Figure: Training System} illustrates the training process for both neural networks. The CNN is trained with image based data, whilst the ANN is trained with feature-based data. Figure \ref{Figure: Execution System} shows a more detailed process of the two-step scheme, in which the processes are further detailed in this paper.

\section{Implementation of the Two-Step Scheme} \label{Section: Methodology}
\subsection{Convolutional Neural Network}
The purpose of Convolutional Neural Network (CNN) implemented in the following two-step scheme is to perform a binary classification on input images to reduce the number of images processed by the ANN. Images with surface damages (cracks) are considered positive, whereas images without surface damage are considered negative. Only images that are classified as positive are passed to the ANN for further processing to highlight the surface damage.

\subsubsection{Image Labelling}  
The CNN requires supervised learning in order to classify positive from negative surface damage. Therefore, the neural network must be trained with images that are clearly labelled either positive or negative. Since a binary classification is required, positive is assigned to a value of `1' and negative is assigned to the value of `0'. The labelled images are used for both training and testing the CNN. 

\subsubsection{CNN Training}
\begin{figure}[h!]
	\centering	
	\includegraphics[scale=1.1]{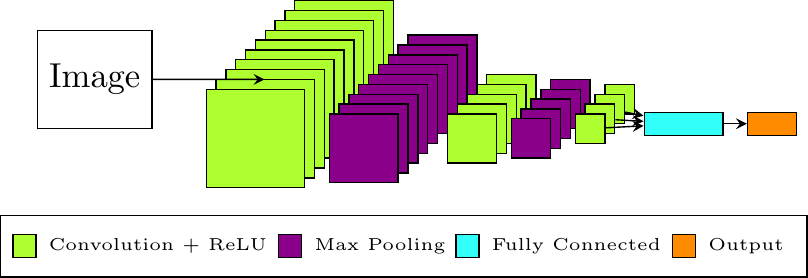}
	\caption{Diagram illustrating the architecture of the Convolutional Neural Network}
	\label{Figure: CNN}
\end{figure}

The structure of the CNN consists of 3 distinct hidden layers, namely, convolutional, max pooling, and fully-connected layer. The architecture of the CNN is shown in Figure \ref{Figure: CNN}. There is a total of 8 layers in the CNN architecture. The CNN accepts colour image inputs in the shape of $227 \times 227 \times 3$. The kernel size for the first convolutional layer is ($198 \times 198$) to produce an output shape of $30 \times 30 \times 227$. The $2 \times 2$ max pooling layer further condenses the input to produce a shape of $15 \times 15 \times 227$. The consecutive layers further reduce the input data to produce a summary vector which categorizes the image into positive or negative. The CNN is trained with a total of 1,260 images, composed of 630 positive and negative surface damage images. Additionally, the CNN is trained with 10 epochs, to ensure that the CNN obtains high accuracy without over-fitting to the neural network to the training images.

\subsection{Artificial Neural Network Training}
The purpose of the Artificial Neural Network (ANN) in the two-step scheme is to highlight positive edges that are considered surface damage (cracks) within the image. The adoption of the ANN is inspired by the schemes proposed in \cite{yang2019long} and \cite{yang_long-bone_2019}. The ANN accepts images that are classified as positive by the CNN. Edges that are cracks within the image are considered positive, whereas all other edges are negative. Both positive and negative edges are present in a positive image. The inputs into the ANN are features, which are extracted from the detected edges within the image.
\subsubsection{Image Preparation Enhancement}
\begin{figure}[h!]
	\centering     
	\subfigure[Positive Damage Image]{\label{fig:enhanced_a}\includegraphics[width=0.22\textwidth]{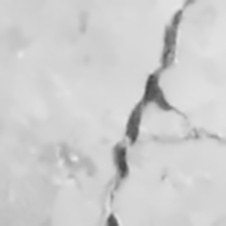}}
	\subfigure[Negative Damage Image]{\label{fig:enhanced_b}\includegraphics[width=0.22\textwidth]{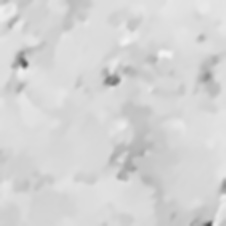}}
	\caption{Edge Detected Images of Positive and Negative Surface Damage}
	\label{Figure: Enhanced Image}
\end{figure}

The images are prepared for edge detection and consequently, feature extraction. The image preparation process converts an image from Red, Green, Blue (RGB) to greyscale. The RGB pixels can be ignored as it does not hold the crucial information for the classification of positive and negative edges within the image. The RGB pixels are replaced with monochrome pixels with values ranging from 0 to 255. The image is further enhanced using tools from the OpenCV library. The raw image is enhanced by applying equalisation, gamma correction, de-noising, and unsharp masking. The enhanced techniques emphasises the image pixel intensity and contrast in preparation for the edge detection. The result of the enhancement is illustrated in Figure \ref{Figure: Enhanced Image} for both positive and negative surface damage.

\subsubsection{Edge Detection}
\begin{figure}[h!]
	\centering     
	\subfigure[Positive Damage Image]{\label{fig:a}\includegraphics[width=0.22\textwidth]{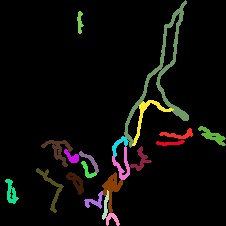}}
	\subfigure[Negative Damage Image]{\label{fig:b}\includegraphics[width=0.22\textwidth]{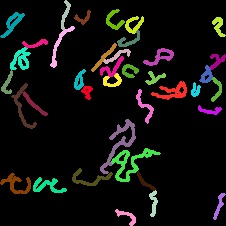}}
	\caption{Edge Detected Images of Positive and Negative Surface Damage}
	\label{Figure: Edge Detection}
\end{figure}

\textit{Canny Edge Detection} is applied to the enhanced image in which the it generates a sequence of points that outlines the edges within the image. Figure \ref{Figure: Edge Detection} illustrates the detected edges for the positive and negative images for the Figures \ref{fig:enhanced_a} and \ref{fig:enhanced_b}, respectively.

\subsubsection{Feature Extraction}
Features are extracted from the detected edges. The edges are constructed with a series of points. The features are extracted based on the points and the angles found within the detected edges \cite{yang_long-bone_2019}, which are shown in Table \ref{Table:Extracted Features}. The first feature, \textit{Number of Edge Points} is the number of points within the detected edge; it encapsulates the size of the edge. The starting points, ($x_1$, $y_1$) and ending points, ($x_2$, $y_2$) are also considered as features. The distance, $d$ feature is calculated using Equation \eqref{eq: distance}. It makes use of the $x$- and $y$-values from the starting and ending points. The gradient feature (given in radians) is determined using Equation \eqref{eq: gradient}. There are four distinct gradient values (in degrees) detected from the edges, namely  $0^{\circ}$, $45^{\circ}$, $90^{\circ}$, and $135^{\circ}$. This is a
result of the Canny Edge Detection function from the OpenCV library \cite{OpenCVWebsite}. The \textit{x-Midpoint} and \textit{y-Midpoint} features are calculated using Equations \eqref{eq: xMidpoint} and \eqref{eq: yMidpoint}, respectively, where $i$ is the indexing of each point within the edge. The remaining features (features 12 to 19) are derived by performing a frequency analysis on the presence of $0^{\circ}$, $45^\circ$, $90^\circ$ and $135^\circ$ from the gradients obtained for \textit{1st Average gradient} and \textit{2nd Average gradient}. \textit{1st Average gradient}, $g^1_avg$ is determined using the algorithm shown in Algorithm \ref{alg: gradient algorithm}. \textit{2nd Average gradient}, $g^2_avg$ is determined by finding the difference between the adjacent gradients in \textit{1st Average gradient}.

\begin{equation}
d = \sqrt{(x_2 - x_1)^2 + (y_2 - y_1)^2}
\label{eq: distance}
\end{equation}

\begin{align}
g = 
\begin{cases}
tan^{-1} \big( \frac{x_2 - x_1}{y_2 - y_1} \big),  & \text{if } y_2 > y_1 \\
tan^{-1} \big( \frac{x_1 - x_2}{y_1 - y_2} \big), & \text{Otherwise} 	
\end{cases}
\label{eq: gradient}
\end{align} 

\textit{1st Average gradient}, $g_{avg}^1$ is determined using the algorithm shown in Algorithm \ref{alg: gradient algorithm}. 
\begin{algorithm}[h]
	\caption{Algorithm for determining the gradient values in degrees}
	\label{alg: gradient algorithm}
	\SetAlgoLined 
	\KwData{gradient value (in radians)}
	\KwResult{gradient value (in degrees)}
	\eIf{$x_i \neq x_{i-1}$}{$g_{temp} = tan^{-1}\frac{y_i - y_{i-1}}{x_i - x_{i-1}}$\\
	\eIf{$g_temp < 0$}{$g = 90^{\circ} + |g_{temp}|$}{$g = g_{temp}$}}		
	{$g = 90^{\circ}$} 
	\Return $g$
\end{algorithm}

\begin{align}
\label{eq: xMidpoint}
M_x &= \frac{\sum_{i = 0}^{N} (\frac{current_{ix} + previous_{ix}}{2})}{N - 1} \\
\label{eq: yMidpoint}
M_y &= \frac{\sum_{i = 0}^{N} (\frac{current_{iy} + previous_{iy}}{2})}{N - 1}
\end{align}

\begin{table}[h]
	\centering
	\caption{Extracted Features and its respective variable types}\label{Table:Extracted Features}
	\begin{tabular}{ccc}
		\hline
		& \textbf{Extracted Feature} & \textbf{Variable} \\
		\hline
		1 & numberOfEdgePoints & $N_p$ \\ 
		2 & $x_1$  & $x_1$\\
		3 & $y_1$  & $y_1$\\
		4 & $x_2$  & $x_2$\\
		5 & $y_2$  & $y_2$\\
		6 & distance  & $d$\\
		7 & gradient  & $g$\\
		8 & 1st-Average-gradient & $g_{avg}^{1}$\\
		9 & 2nd-Average-gradient & $g_{avg}^{2}$ \\
		10 & x-Midpoint & $M_x$\\
		11 & y-Midpoint & $M_y$\\
		12 & number-Of-Zero-Gradients & $N_{g_{0}}$\\
		13 & number-Of-45-Gradients & $N_{g_{45}}$ \\
		14 & number-Of-90-Gradients & $N_{g_{90}}$\\
		15 & number-Of-135-Gradients & $N_{g_{135}}$\\
		16 & number-Of-Zero-Difference-Gradients & $N_{\Delta g_{0}}$\\
		17 & number-Of-45-Difference-Gradients & $N_{\Delta g_{45}}$ \\
		18 & number-Of-90-Difference-Gradients & $N_{\Delta g_{90}}$\\
		19 & number-Of-135-Difference-Gradients & $N_{\Delta g_{135}}$\\
		\hline
	\end{tabular}
\end{table}

\subsubsection{Principle Component Analysis}
Principle Component Analysis (PCA) is an analysis tool most commonly used for dimensionality reduction \cite{WOLD198737, Jolliffe2016}. The analysis is performed by determining the principle components that holds the most
varying information through deriving the co-variance matrix, eigenvectors, and eigenvalues. The detailed steps of PCA is described below:
\begin{enumerate}
	\item A normalised matrix, $M_{norm}$ of the features is used to determine the covariance matrix, $C$, which is expressed in Equation \eqref{eq: covariance matrix}
	\begin{equation}
		\label{eq: covariance matrix}
		C = \frac{M^T_{norm} \times M_{norm}}{m - 1}
	\end{equation}
	where $m$ is the number of edges. 
	\item The eigenvectors $v_1, v_2, v_3, ..., v_i$ corresponding to the eigenvalues $\lambda_1, \lambda_2, \lambda_3, ..., \lambda_i$ are calculated. Assuming that the eigenvalues are in descending order such that: $\lambda_1 \geq \lambda_2 \geq \lambda_3 \geq, ..., \geq \lambda_i$ is the first principal component and $v_1$ consists of the main characteristics for the given data.
	\item The principal components, Pi relate to the dimensions of the edges within the image can be expressed as follows:
	\begin{equation}
		P_i = \lambda_{i1}Z_1 + \lambda_{i2}Z_2 + ... + \lambda_{im}Z_m
	\end{equation}
	where $Z_m$ represents the features extracted from the edges.
\end{enumerate}
PCA is employed to evaluate the extracted features in order to determine the feature that holds the distinguishing information for the classification of positive and negative edge \cite{yang2019long}. The results obtained through the use of PCA for positive and negative edge features are illustrated in Figure 7. The dominant features for a positive edge is \textit{Number of 135 Difference Gradients and Gradient}.

\begin{figure}[ht!]
	\label{fig: pca}
	\includegraphics[scale=0.45]{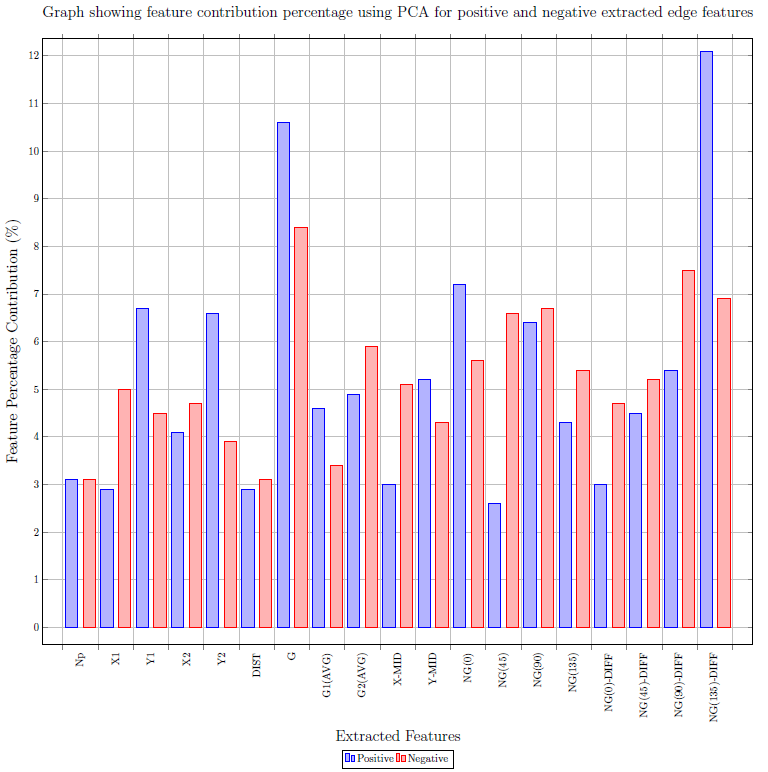}
	\caption{Graph illustrating the contribution for each extracted feature for a
		positive and negative edge.}
\end{figure}

\subsubsection{Feature Labelling}
Each set of the extracted feature represents an edge. The labelling of each set of extracted features indicates whether the edge is surface damage (crack) or not. The labelling process is performed manually through a Graphical User Interface (GUI). The user utilises an area selection tool for the labelling process. Edges that fall within the selected area are labelled positive edge damages, whilst all other edges are negative. Both positive and negative edges are present within a image that is classified as positive surface damage by the CNN.

\subsubsection{Artificial Neural Network Training} \label{Section: Neural Network}
\begin{figure}[ht!]
	\begin{center}
		\includegraphics[scale=0.62]{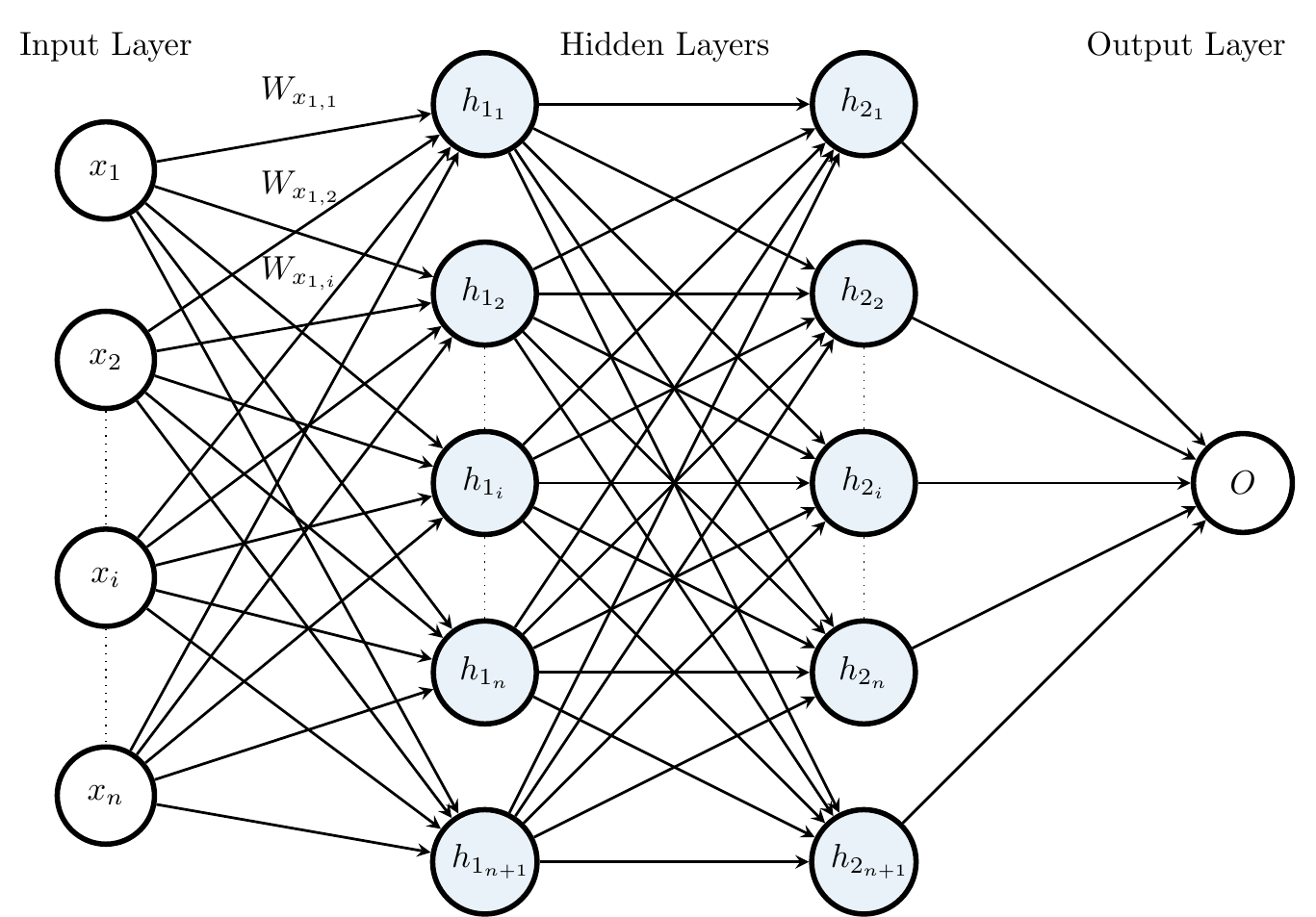}
	\end{center}
	\caption{Diagram showing a graphical set-up of the Artificial Neural Network}
	\label{fig:neural network}
\end{figure}

The features extracted from each edge are used as inputs into the ANN. The ANN consists of 3 distinct layers, namely, input, hidden, and output layer. The layers are connected through weighted connections between each node. The input layer consists of 19 nodes for the 19 extracted features. The hidden layer consists of 20 nodes, whilst there is one node in the output layer. The ANN performs a binary classification to determine whether the given edge is positive or negative surface damage. A graphical representation of the ANN architecture is shown in Figure \ref{fig:neural network}. The ANN is trained with a total of 376 positive and negative edges from 57 images. The training of the ANN is less computationally intensive compared to the training of the CNN due to the reduced number of dimensionality, as the inputs to the ANN are the extracted features.

\section{Results and Analysis} \label{Section: Results and Analysis}
\subsection{CNN vs ANN}
\begin{figure}[h!]
	\centering	
	\includegraphics[scale=0.7]{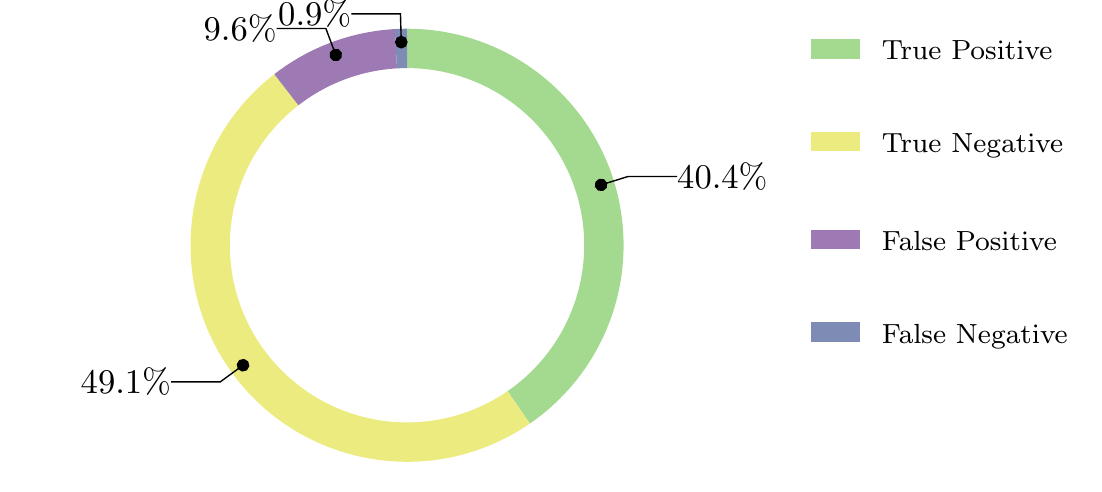}
	\caption{Pie chart illustrating the detected True Positive, True Negative, False Positive, and False Negative for the CNN surface damage classification}
	\label{Figure: CNN_TrueFalse}
\end{figure}

\begin{figure}[h!]
	\centering	
	\includegraphics[scale=0.7]{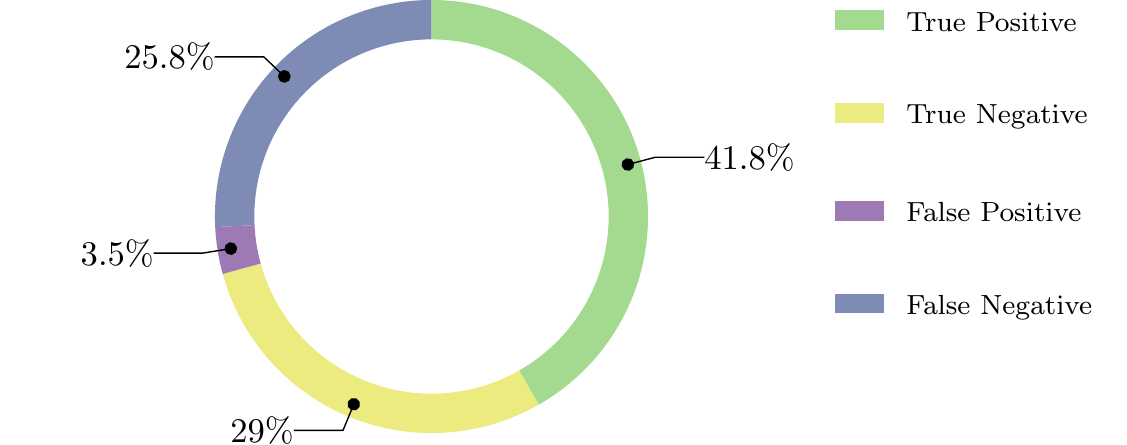}
	\caption{Pie chart illustrating the detected True Positive, True Negative, False Positive, and False Negative ANN edge classification}
	\label{Figure: ANN_TrueFalse}
\end{figure}

Both CNN and ANN are evaluated individually to determine its individual performance. The detection accuracy for both CNN and ANN are illustrated in Figures \ref{Figure: CNN_TrueFalse} and \ref{Figure: ANN_TrueFalse}, respectively. The results reflect the true positive, $TP$, true negative, $TN$,
false positive, $FP$, and false negative, $FN$ detections. The CNN is tested with a total of 540 images that is comprised of an equal number of positive and negative images. The ANN is tested with 500 edges, where 300 are positive and 200 are negative edges. The accuracy of the neural networks are calculated using equation \eqref{eq: accuracy}. The accuracy for CNN is 89.4\%, whilst the ANN accuracy is 70.7\%. The CNN has a higher accuracy compared to the ANN. The detection of the false negative is crucial as it is indicative of the incorrect detection where fault is present but classified as absent. The false negative detection for CNN is 0.9\%, whilst for the ANN the false negative detection is 25.8\%. The ANN has a far greater false negative detection, which contributes to lowering the detection accuracy of the overall ANN.

\begin{equation}
	\text{Accuracy} = \frac{TP + TN}{TP + TN + FP + FN}
	\label{eq: accuracy}
\end{equation}

\subsection{Performance of Two-Step Scheme}
\begin{figure}[h!]
	\centering	
	\includegraphics[scale=0.7]{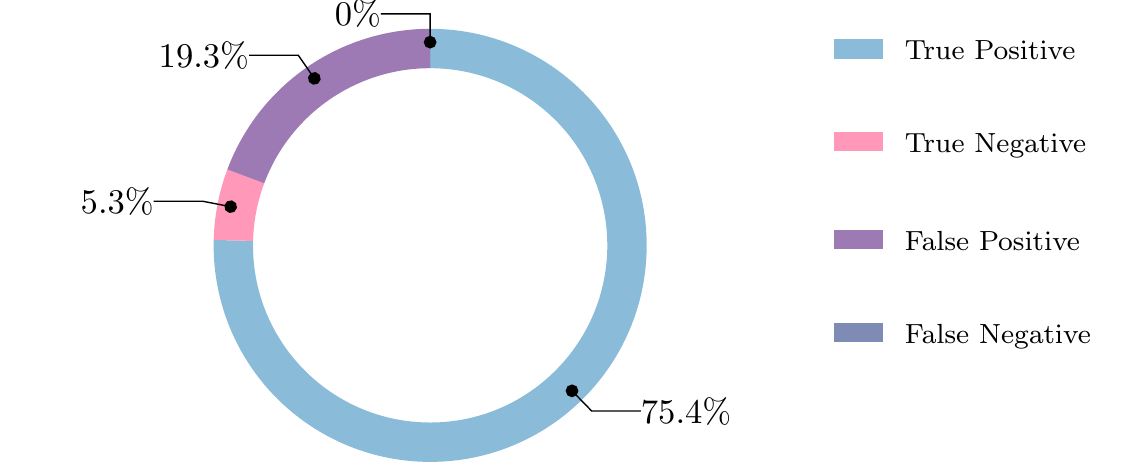}
	\caption{Pie chart illustrating the detected True Positive, True Negative, False Positive, and False Negative surface damage classification}
	\label{Figure: Combined CNN_TrueFalse}
\end{figure}

\begin{figure}[h!]
	\centering	
	\includegraphics[scale=0.7]{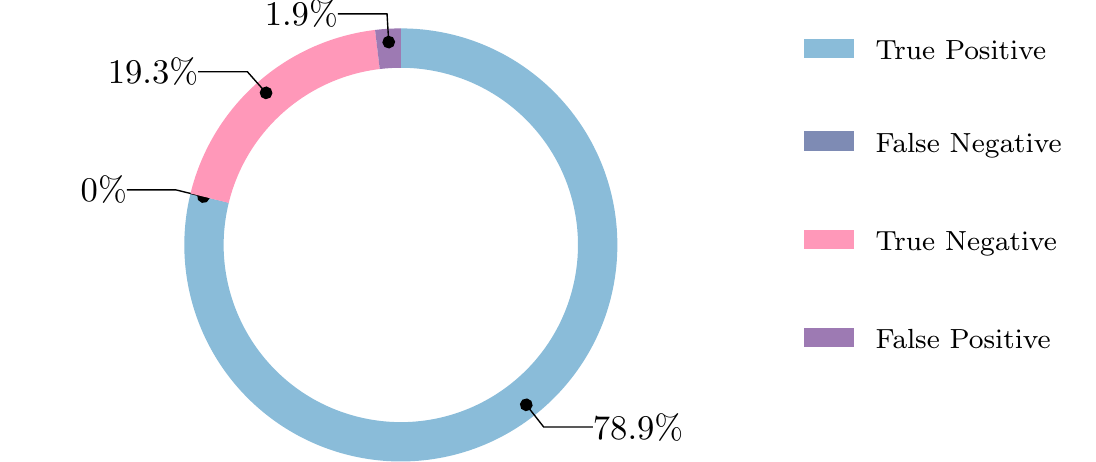}
	\caption{Pie chart illustrating the detected True Positive, True Negative, False Positive, and False Negative surface damage classification}
	\label{Figure: Combined ANN_TrueFalse}
\end{figure}

In the two-step scheme, the CNN focuses on the image-based classification whilst the ANN focuses on feature-based classification and as a result highlights the damaged area within the image. The weakness of the CNN compared to the ANN is that the CNN is unable to highlight the damage in the image, whereas the weakness of the ANN is that it presents a higher false negative detection compared to the CNN. The combination of both neural networks compensates for one another’s weaknesses by combining its individual strengths. This improves the detection of the surface damage. CNN is assigned to separate positive surface damage images from negative images. Only the positive images are further processed by the ANN. The ANN highlights the positive edges that are considered surface damage within the image. For the two-step scheme, the accuracy of the CNN is 80.7\%, and the accuracy of the ANN is 98.1\%. The detections for both neural networks are illustrated in Figures \ref{Figure: Combined CNN_TrueFalse} and \ref{Figure: Combined ANN_TrueFalse}. The accuracy of the CNN is 80.7\%, which is less than the accuracy of the individual CNN evaluation. This is due to the false positive detection in the given testing sample. False positives are tolerated as it is an indication of a cautious system. The crucial detection is false negative detection, in which both CNN and ANN has 0\% false negative detection for the two-step scheme.

\begin{figure}[h!]
	\centering	
	\includegraphics[scale=0.45]{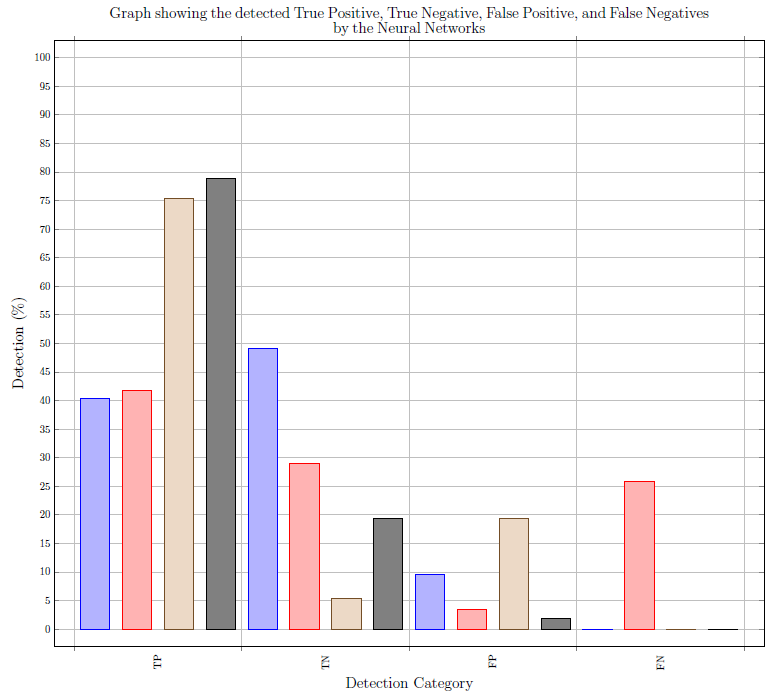}
	\caption{Graph illustrating the detected True Positive, True Negative, False Positive, and False Negative surface damage classification}
	\label{Figure: TP detection}
\end{figure}

\begin{figure}[h!]
	\centering	
	\includegraphics[scale=1.5]{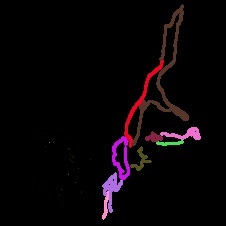}
	\caption{Image illustrating the ANN detected positive edges for Figure \ref{fig:enhanced_a}}
	\label{Figure: contour}
\end{figure}

Figure \ref{Figure: TP detection} illustrates the comparison of the detection between the individual neural networks and the neural networks in the two-step scheme. The detection of the false positive and
false negative detections are crucial as this is not reflected in the accuracy of the neural networks. The false negative detection is the most crucial, as false negatives classify positive edges as negative edges. Therefore the false negative indicates the percentage of positive edges that go undetected by the neural networks. The results obtained from the two-step ANN show that the false positive and false negative is decreased to 0\%, compared to the false positive and false negative detections for the individual neural networks. This is an improvement from the individual ANN false negative detection of 25.8\%. The result of the produced by the ANN is shown in Figure \ref{Figure: contour}.

\section{Future Improvements} \label{Section: Future Improvements}
Future improvements to the two-step scheme includes decreasing the kernel size of the CNN for the first convolutional layer, such that finer automated feature extraction can be performed. This will increase the performance of the CNN, resulting in a higher accuracy. However, this improvement requires more computational power. Future improvements for the ANN, includes the labelling of the training data, as the
labelling is performed using area selection. This means that edges that are not considered damage edges are labelled positive. This alters the training weights in the neural network. Hence, it decreases the overall accuracy of the ANN, as well as increasing false detections.

\section{Conclusion}
To conclude, this paper proposes a two-step surface damage detection scheme. The scheme consists of a CNN and an ANN. The objective of the CNN is to classify images as positive or negative. An image is considered positive when there is surface damage (crack) present in the image, otherwise it is considered negative. The objective of the ANN is to highlight the edges of the surface damage within the image. The two-step scheme feeds image inputs into the CNN to classify the image. The images classified with positive surface damage is passed to the ANN, whereas all other images are not further considered. The positive images are further pre-processed for edge detection, which is then followed by feature extraction. A total of 19 features are extracted from the detected edges within the image. These features are fed into the ANN as inputs to determine whether the edge is positive or negative. A positive image contains both positive and negative edges. Hence the ANN highlights only the positive edges that are considered as surface damage. PCA is applied to the extracted features, the dominant features for an edge to be considered positive are \textit{Number of 135 Gradient Difference} and the \textit{Gradient}
feature. The CNN achieved an 80.7\% accuracy, whilst the ANN achieved 98.1\% detection accuracy. The false negative detection for both neural networks are 0\%. This indicates that the system is not falsely allowing positive images and edges to be ignored.

\bibliographystyle{IEEEtran}
\bibliography{reference}

\end{document}